\PassOptionsToPackage{table}{xcolor}

\documentclass[sigconf]{acmart}

\AtBeginDocument{%
  \providecommand\BibTeX{{%
    \normalfont B\kern-0.5em{\scshape i\kern-0.25em b}\kern-0.8em\TeX}}}

\usepackage[table]{xcolor}
\usepackage[]{algorithm}
\usepackage{algpseudocode}
\usepackage{multirow}
\usepackage{tabularx}
\usepackage{array}
\usepackage{graphicx}
\usepackage{rotating}
\usepackage{hyperref}
\usepackage{subfigure}
\usepackage{csquotes}
\usepackage{pgfplots, pgfplotstable}
\usepackage{bbm}
\usepackage{amsmath}
\usepackage{float}
\usepackage{placeins}
\usepackage{stmaryrd,scalerel}
\usepackage{pifont}
\newcommand{\cmark}{\ding{51}}%
\newcommand{\xmark}{\ding{55}}%

\pgfplotsset{compat=1.8}

\usepackage{tikz}
\usetikzlibrary{arrows.meta,
                chains,
                positioning,
                intersections,
                shapes.geometric,
                fit
                }

\newcommand{\algorithmname}{Algorithm}
\def\ALG@name{\algorithmname}
\providecommand*{\ALG@name}{Algorithm}

\definecolor{TableHighlight}{gray}{0.9}

\begin{document}
\settopmatter{printacmref=false} 
\renewcommand\footnotetextcopyrightpermission[1]{}
\pagestyle{plain}

\title{A Systematic Review of Conformal Inference Procedures for Treatment Effect Estimation: Methods and Challenges}

\author{Pascal Memmesheimer}
\orcid{0009-0004-5775-1005}
\affiliation{%
  \institution{Engineering Mathematics and Computing Lab\\ Heidelberg University}
  \streetaddress{Im Neuenheimer Feld 205}
  \city{Heidelberg}
  \state{Baden-Württemberg}
  \postcode{69120}
  \country{Germany}}
\additionalaffiliation{Heidelberg Institute for Theoretical
Studies, Data Mining and Uncertainty Quantification Group}
\additionalaffiliation{Medical Faculty Mannheim of Heidelberg
University, Data Analysis and Modeling in Medicine}
\email{pascal.memmesheimer@uni-heidelberg.de}

\author{Vincent Heuveline}
\orcid{0000-0002-2217-7558}
\affiliation{%
	\institution{Engineering Mathematics and Computing Lab\\ Heidelberg University}
	\streetaddress{Im Neuenheimer Feld 330}
	\city{Heidelberg}
	\state{Baden-Württemberg}
	\postcode{69120}
	\country{Germany}}
\authornotemark[1]
\email{vincent.heuveline@uni-heidelberg.de}

\author{Jürgen Hesser}
\orcid{0000-0002-4001-1164}
\affiliation{%
  \institution{Data Analysis and Modeling in Medicine\\ Heidelberg University}
  \streetaddress{Theodor-Kutzer-Ufer 1-3}
  \city{Mannheim}
  \state{Baden-Württemberg}
  \postcode{68167}
  \country{Germany}}
\email{juergen.hesser@medma.uni-heidelberg.de}

\begin{abstract}
Treatment effect estimation is essential for informed decision-making in many fields such as healthcare, economics, and public policy. 
While flexible machine learning models have been widely applied for estimating heterogeneous treatment effects, quantifying the inherent uncertainty of their point predictions remains an issue.
Recent advancements in conformal prediction address this limitation by allowing for inexpensive computation, as well as distribution shifts, while still providing frequentist, finite-sample coverage guarantees under minimal assumptions for any point-predictor model. 
This advancement holds significant potential for improving decision-making in especially high-stakes environments. 
In this work, we perform a systematic review regarding conformal prediction methods for treatment effect estimation and provide for both the necessary theoretical background.
Through a systematic filtering process, we select and analyze eleven key papers, identifying and describing current state-of-the-art methods in this area.
Based on our findings, we propose directions for future research.
\end{abstract}


\maketitle

\section{Introduction}
Uncertainty quantification is crucial for high-stakes decision-making, especially in the medical domain.
Here, it is essential to establish robustness and trustworthiness of models.
In the past, there has been widespread adoption of black-box models like neural networks or other difficult-to-interpret models like gradient-boosted trees that often only provide point predictions.
The reliance on such point predictions is insufficient for assessing risks for reliable, robust, and trustworthy decision-making.

Conformal prediction promises remedy.
This framework of methods conformalizes the outputs of any point predictor, s.t. marginally valid prediction regions can be computed.
Recently, these methods have become popular which can be seen by the sheer number of papers published on this topic \cite{10.5555/2349018}.
The application of conformal prediction for causal inference is not trivial, however, since conformal prediction relies on the notion of \textit{exchangeability} which is inherently violated in causal inference since the counterfactual is never observed, thus, the inference problem involves a potential covariate shift between the target distribution and the sampling distribution \cite{lei_conformal_2021}.
The recent introduction of \textit{weighted exchangeability} has intensified interest in conformal prediction for treatment effect estimation, as it allows for handling distributional shifts \cite{tibshirani2020conformalpredictioncovariateshift}.

\subsection{Our Contribution}
In this systematic review, we provide a comprehensive overview of conformal prediction techniques for treatment effect estimation. 
By organizing, filtering, and summarizing various approaches, we highlight their suitability for different types of problems.
To the best of our knowledge, no other review paper exists for applying conformal prediction for treatment effect estimation.

We perform a comprehensive, structured search on all research papers from 2005 until 2025 using multiple electronic databases.
For our filtration process, we establish both inclusion criteria and quality standards to ensure a transparent and well-structured approach.
Through this process, we identify eleven papers, which we discuss in detail and use to answer our research questions.
This allows interested researchers to join this important research area.

This paper is organized as follows.
In \autoref{sec:preliminaries}, we review the foundations of conformal prediction and discuss key concepts and definitions in causal inference, including popular methods for estimating treatment effects.
Thereafter, we discuss our methodology for the search and filtration process in \autoref{sec:methodology}.
In \autoref{sec:results_total} we summarize various methods of conformal inference for estimating treatment effects and analyze the meta-data of the papers published.
We discuss our findings and possible future research directions in \autoref{sec:discussion}.
Conclusions are given in \autoref{sec:conclusion}.

\section{Preliminaries}
\label{sec:preliminaries}
In the following two sections, we introduce the theoretical foundations of the relevant conformal prediction methods for this work.
Thereafter, we introduce important causal inference terms as well as popular methods for estimating treatment effects.

\subsection{Conformal Prediction}
Conformal prediction, also known as conformal inference, is a method originally introduced by Vladimir Vovk et al. \cite{vovk_algorithmic_2005, papadopoulos_inductive_2002, lei_distribution-free_2013}, and has since been refined over many years. 
It has garnered significant attention in both the statistics and machine learning communities for its ability to provide distribution- and model-free finite-sample uncertainty quantification in the form of statistically valid \textit{prediction regions}.
The method can be used for classification tasks for which it generates \textit{prediction sets} (e.g., $\{\text{cat}, \text{dog}, \text{crocodile}\}$).
In regression tasks, \textit{prediction intervals} are generated (e.g., $[1000, 5000]$).
We will refer in both cases to the more general nomenclature \textit{prediction region} and specify only in more detail when needed.

There are many variants of conformal prediction that have been introduced and refined in the past two decades.
For a more rigorous definition and introduction, we refer to \citet{theoretical}, \citet{Fontana_2023, angelopoulos2022gentleintroductionconformalprediction} and especially the comprehensive book by \citet{vovk_algorithmic_2005}.
We will focus on the predominant method of \textit{split conformal prediction} since it decreases the heavy computational burden of the original method, \textit{full conformal prediction}.
Other variants have been introduced, notably variants that make use of cross-validation.
For this, we refer to \citet{barber2020predictiveinferencejackknife, vovk_cross_conformal_2015}, as well as to \citet{vovk_nonparametric2017} who also describe a procedure for calculating conformal predictive distributions. 

In \textit{split conformal prediction}, the data is divided into three subsets: training, calibration, and test, as opposed to the usual train/test division.
It then uses the aforementioned training data set to fit an arbitrary learner $\hat{f}: \mathbb{R}^d\rightarrow \mathbb{R}$, i.e., producing one-dimensional point estimates from $d$-dimensional inputs, in a first step.
In a second step, the learners predictions are being \textit{conformalized} (or calibrated), s.t. we can make statistically valid statements of our models predictions with only minimal guarantees which we will elaborate later on.
The procedure guarantees \textit{marginal coverage} regardless of the learner’s predictive accuracy, though the resulting regions may be wide.
The goal is therefore to construct non-trivial prediction regions that both attain the desired marginal coverage and remain as tight as possible.

Let $(X_1, Y_1), \dots, (X_m, Y_m)$ be a sequence of random variables sampled i.i.d.\ from some distribution $\mathscr{P}_{XY}$ on $\mathbf{X}\times \mathbf{Y}$, with marginals $\mathscr{P}_X$ and $\mathscr{P}_Y$. 
Each pair is called an \textit{example}, $Z_i := (X_i, Y_i) \in \mathbf{Z} := \mathbf{X} \times \mathbf{Y}$ for $i = 1, \dots, m$. 
Here, $X_i \in \mathbf{X}$ represents the features and $Y_i \in \mathbf{Y}$ the response.
Throughout this work, we focus on the regression setting and assume without loss of generality that $\mathbf{Y} = \mathbb{R}$ and $\mathbf{X} = \mathbb{R}^d$, where $d$ denotes the dimensionality of the feature space.
We denote by lowercase letters realizations of random variables.

\begin{figure}
    \centering
    \includegraphics[width=\linewidth]{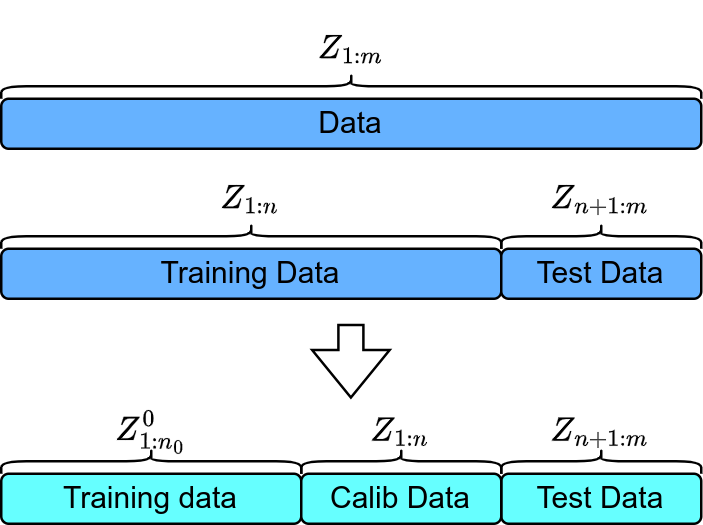}
    \caption{Usual data split (top) and data split for split conformal prediction (bottom).}
    \label{fig:datasplit}
\end{figure}

For both classification and regression, the data is split (see \autoref{fig:datasplit}) into three distinct multisets:

\begin{enumerate}
    \item A \textit{training set} $Z_{\text{train}} := Z^0_{1:n_0} = \{Z^0_1, \dots, Z^0_{n_0}\}$
    \item A \textit{calibration set} $Z_{\text{calib}} := Z_{1:n} = \{Z_1, \dots, Z_n\}$
    \item A \textit{test set} $Z_{\text{test}} := Z_{n+1:m} = \{Z_{n+1}, \dots, Z_m\}$
\end{enumerate}

A central notion of conformal prediction is the \textit{nonconformity measure}.
Informally, a nonconformity measure estimates how "unusual" a sample looks with respect to the population of data points.
A nonconformity measure is a function $A: \mathbf{Z^{(*)}} \times \mathbf{Z} \rightarrow \overline{\mathbb{R}}$, where $\mathbf{Z^{(*)}}$ is a finite multiset of elements in $\mathbf{Z}$, and $\overline{\mathbb{R}}$ is the extended real numbers, i.e., $\overline{\mathbb{R}} = \mathbb{R} \cup \{-\infty, \infty\}$.
A nonconformity measure induces a \textit{nonconformity score} at each data point:

\[
R_i := A((X_i, Y_i); Z_{\text{calib}}), \quad i=1,\dots,n.
\]

In regression problems, a common choice for $A$ is the absolute residual defined as $A((x,y); Z_{\text{calib}}) :=  |\hat{f}(x;  Z_{\text{train}}) - y|$, where $\hat{f}(x)$ is a point prediction made by a learner $\hat{f}$ previously fit on the training set $Z_\text{train}$.

The nonconformity measure in split conformal prediction reduces to $A((x,y); Z_{\text{calib}}) = A(x,y)$  since the nonconformity measure does not depend on the calibration fold.
The choice of the nonconformity measure is subject to the individual learner as well as the problem at hand and remains an active area of research \cite{kato2023}.

In the split conformal prediction procedure, given random variables $(X_i, Y_i) \in Z_{\text{calib}}$ (see \autoref{fig:datasplit}), their realizations $(x_i, y_i)$, a learner $\hat{f}$ fitted on the training data $Z_{\text{train}}$, and a nonconformity measure $A$, we compute \textit{nonconformity scores} by

\[    
r_i:=  A(x_i, y_i) =|\hat{f}(x_i) - y_i|, \quad i=1,\dots,n.
\]

Based on the ordered list of nonconformity scores, we can compute the conformal quantile by 
\[
\hat{q}_\alpha=\text{Quantile}_{(1-\alpha)(1+\frac{1}{n})}(r_1, \dots, r_n)\footnote{The factor $(1-\alpha)(1+\tfrac{1}{n})$ accounts for finite-sample calibration.},
\]
where the $\text{Quantile}_\alpha$ is defined as $\text{Quantile}_\alpha(z)=\text{inf}\left\{v: \widehat{F}_z(v) \geq \alpha \right\}$ and $\widehat{F}_z$ is a function $\widehat{F}_z:\mathbb{R}\rightarrow[0,1]$, s.t., $\widehat{F}_z(v)= \frac{1}{n}\sum^n_{i=1}\mathbbm{1}\{z_i \leq v\}$.

Consequently, we can define a conformal prediction interval for a point $x\in \mathbb{R}^d$ by:

\begin{equation}
\label{eq:predictioninterval}
    \widehat{C}_n(x) = [\,\hat f(x) - \hat q_\alpha,\; \hat f(x) + \hat q_\alpha].
\end{equation}

Under the assumption that the examples $Z_1, \dots, Z_{n+1}$ are exchangeable (and thus, also the nonconformity scores $R_1, \dots R_n$), the calculated prediction region satisfies 

\begin{equation}
\label{eq:conformal}
    \mathbb{P}(Y_{n+1}  \in \widehat{C}(X_{n+1})) \geq 1-\alpha,
\end{equation}

where the probability is taken over the $n+1$ points and $\alpha$ is a user-specified \textit{error rate} or \textit{significance level}.
The property of fulfilling \autoref{eq:conformal} is called \textit{marginal coverage} since the probability is marginal over the randomness in the calibration and test data \cite{angelopoulos2022gentleintroductionconformalprediction}.

Conformal prediction is not only able to conformalize predictions in the regression setting but can also be used for classification.
For neural networks in the classification setting, a possible nonconformity measure is the softmax function which outputs a score per class which can be interpreted as a probability \cite{vovk_algorithmic_2005}, i.e., $A(x_i, y_i) = 1 - \hat{f}(x_i)_{y_{true}}$, where $\hat{f}(x_i)_{y_{true}}$ denotes the score of the model for the true class of the $i$th sample.
As before, we calculate for all of the calibration examples nonconformity scores $r_i = A(x_i, y_i), i=1,\dots,n$.
Thereafter, we compute the empirical quantile \[\hat{q}_\alpha = \text{Quantile}_{(1-\alpha)(1+\frac{1}{n})}(r_1, \dots, r_n).
\]
Then, we can construct the prediction set as 
\begin{equation}
\label{eq:predictionintervalclassification}
    \widehat{C}_n(x)= \{y \in \mathbf{Y}: \hat{f}(x)_y \geq 1-\hat{q}_{\alpha}\},
\end{equation}
where $\hat{f}(x)_y$ denotes the score for a specific label. 
Thus, we include every class in the prediction set is that at least as conforming as the calculated quantile $1-\hat{q}_{\alpha}$.
The constructed sets satisfy \autoref{eq:conformal}.

\begin{algorithm}
\caption{Split Conformal Prediction: General Case}
\label{alg:conformal}
\begin{algorithmic}[1]
\Require Examples $(z_1,, \dots, z_m) \in \mathbf{Z}$, significance level $\alpha$, learner $\hat{f}$, nonconformity measure $A$, $r = \emptyset$.
\State Split examples into a proper \textit{training set} $Z_{\text{train}} = \{z^0_1, \dots, z^0_{n_0}\}$, a proper \textit{calibration set} $Z_{\text{calib}}=\{z_{1}, \dots, z_n\}$ and in a proper \textit{test set} $Z_{\text{test}}=\{z_{n+1}, \dots, z_m\}$.
\State Fit an arbitrary learner $\hat{f}$ on $Z_{\text{train}}$.
\State Define a heuristic notion of uncertainty that defines how "unusual" an example looks to previous examples called the nonconformity measure $A(x,y)$.
\State For each element $z_i \in Z_{\text{calib}}$, calculate the nonconformity score $r_i=A(x_i, y_i)$ with $i=1,\dots,n$ and add it to the set $r$. Afterwards, order the nonconformity scores in the set $r$, s.t. $r_{(1)} \leq r_{(2)} \leq \dots \leq r_{(n)}$.
\State Compute the quantile of the nonconformity scores by $\hat{q}_\alpha = Q_{(1-\alpha)(1+\frac{1}{n})}(r)$.
\State \Return For each $z_i \in Z_{\text{test}}$ calculate the prediction region $\widehat{C}_n(x)$ using the previously computed quantile $\hat{q}_\alpha$ and \autoref{eq:predictioninterval} for regression tasks, or \autoref{eq:predictionintervalclassification} for classification tasks.
\end{algorithmic}
\end{algorithm}

The algorithmic steps of computing intervals or sets using split conformal prediction is shown in \autoref{alg:conformal}.
The conformal prediction procedure achieves exact valid coverage (i.e., the errors do not exceed a specified error level $\alpha$ as shown in \autoref{eq:conformal}) with minimal assumptions, without assuming any specific underlying data distribution, model or model performance, in finite samples.

However, for split conformal prediction, the examples of calibration and test set $Z_{1}, \dots, Z_{m}$ need to be drawn from an exchangeable sequence, which is a slightly weaker assumption than independent and identically distributed data.
A sequence of random variables $U_1, \dots, U_n$ is exchangeable if for any permutation of the sequence, the joint distribution of these random variables is invariant under this permutation, i.e. $U_1, \dots U_n \overset{d}{=} (U_{\pi(1)}, \dots, U_{\pi(n)})$ for any permutation $\pi$.
If this assumption is violated, then \autoref{eq:conformal} does not hold and the coverage can get arbitrarily poor.

One of the key design considerations that one must consider for conformal prediction is \textit{adaptivity} \cite{angelopoulos2022gentleintroductionconformalprediction}. 
\textit{Adaptivity} refers to the property that prediction regions should adapt to the "difficulty" of a prediction problem.
Hence, the prediction region for a sample should be large when the model is unsure of its prediction and small when the model is confident in its prediction.

Adaptivity is usually formalized by asking for \textit{conditional coverage} to be fulfilled \cite{angelopoulos2022gentleintroductionconformalprediction}:

\begin{equation}
    \mathbb{P}(Y_{n+1}  \in \widehat{C}(x) \;| \; X_{n+1}=x) \geq 1-\alpha    
\end{equation}

For each sample $X_{i}$, it would be ideal if the prediction region was guaranteed to contain the true value with probability $1-\alpha$ at a point $x\in \mathbb{R}^d$.
However, without additional assumptions—such as knowledge of the true data-generating process or the imposition of a Bayesian prior—this guarantee is unattainable in finite samples for rich object spaces \cite{barber_limits_2020, lei_distribution-free_2013}.
Nevertheless, assessing whether approximate coverage holds remains an important consideration, especially for specific subgroups in data.

Another critical consideration in conformal prediction is the size of the calibration set. 
While a larger calibration set generally improves reliability, empirical evidence suggests that a sample size of 1,000 often suffices for practical purposes \cite{angelopoulos2022gentleintroductionconformalprediction}. 
In scenarios with limited data availability, cross-conformal prediction provides an alternative by leveraging cross-validation—that is, partitioning the data into multiple folds for repeated calibration. 
A special case of this approach, leave-one-out cross-conformal prediction, is known as the jackknife+ method.
While cross-conformal prediction lacks a provable coverage guarantee \cite{vovk2025}, the jackknife+ method offers a theoretically valid $1-2\alpha$ coverage guarantee under mild assumptions \cite{barber2020predictiveinferencejackknife}.

There have been many recent developments for Conformal Prediction: They are being used in conjunction with Shapley values to explain various types of predictive uncertainty \cite{NEURIPS2023_16e4be78}, and have been adapted to several different domains.
In the following, we will define two methods of conformal prediction that are of special importance for the methods described in \autoref{sec:results_total}.

\subsection{Conformalized Quantile Regression}
Conformalized quantile regression (CQR), introduced by \citet{romano_conformalized_2019}, integrates quantile regression with conformal prediction.
The popularity of this method stems from the inherent advantages of quantile regression, which provides well-calibrated prediction regions even before conformalization. 
Moreover, quantile regression offers asymptotically valid conditional coverage guarantees. 
Through the conformalization process, these favorable properties are preserved, leading to narrower prediction intervals \cite{angelopoulos2022gentleintroductionconformalprediction}.

Quantile regression (often referred to as nonparametric quantile regression) estimates a given quantile of $Y$ conditional on $X$.
The $\alpha$-th conditional quantile function is defined as $q_\alpha(x):=inf\{y \in \mathbb{R}: F(y \;|\; X=x) \geq \alpha\}$ where $F(y\;|\;X=x) := \mathbb{P}(Y \leq y \; | \; X = x)$ is called the conditional distribution function.
Upper and lower quantiles $\alpha_{lo}$ and $\alpha_{hi}$ can be arbitrarily defined and a conditional prediction interval for $Y$ given $X=x$ with miscoverage rate $\alpha$ is given by $C_n(x)=[q_{\alpha_{lo}}(x), q_{\alpha_{hi}}(x)]$, and where $q_{\alpha_{lo}}(x)$, $q_{\alpha_{hi}}(x)$ defines the lower conditional quantile and upper conditional quantile respectively.
The prediction intervals satisfies the conditional coverage

\begin{equation}
\label{eq:coverage}
    \mathbb{P}(Y \in C(X) \; |\; X=x)\geq 1-\alpha
\end{equation} 
by construction \cite{romano_conformalized_2019}.

The apparent solution is easy:
Estimate $q_{\alpha_{lo}}(x)$ and $q_{\alpha_{hi}}(x)$; and for a new test point $X_{n+1}$ construct the prediction region $\hat{C}(X_{n+1})=[q_{\alpha_{lo}}(X_{n+1}), \; q_{\alpha_{hi}}(X_{n+1})]$.
Because the estimation of these intervals can get arbitrarily bad, the coverage statement in \autoref{eq:coverage} is not guaranteed and there are no finite-sample guarantees. 

CQR aims to combine conformal prediction and quantile regression for coverage guarantees.
First, the data is split into a proper training set $Z_{\text{train}}$ and a calibration set $Z_{\text{calib}}$.
Then for any quantile regression learning algorithm $\hat{f}$, the two aforementioned functions are fit such that: $\hat{f}(\{(X_i, Y_i): i \in Z_{\text{train}}\}) \rightarrow \{q_{\alpha_{lo}}, q_{\alpha_{hi}}\}$.
As usual in conformal prediction procedures, conformity scores are computed on the held-out calibration data set to quantify the error of the prediction interval computed before.
\citet{romano_conformalized_2019} define and use the following nonconformity score:
\[
    r_i := max\{q_{\alpha_{lo}}(x_i) -y_i, y_i -q_{\alpha_{hi}}(x_i)\}
\]

This conformity score is defined specifically to account for both undercoverage and overcoverage.
Conformalized prediction intervals are then constructed by 
\begin{multline}
    \widehat{C}(x_{n+1}) = [\hat{q}_{\alpha_{lo}}(x_{n+1})- \text{Quantile}_{(1-\alpha)(1+\frac{1}{n})}(r), \\\hat{q}_{\alpha_{hi}}(x_{n+1}) + \text{Quantile}_{(1-\alpha)(1+\frac{1}{n})}(r)],
\end{multline}
where $\text{Quantile}_{{(1-\alpha)(1+\frac{1}{n})}}(R)$ is defined as before.
These regions are marginally valid and satisfy \autoref{eq:conformal}, given that the data is exchangeable.

\subsection{Weighted Conformal Prediction}
\citet{tibshirani2020conformalpredictioncovariateshift} introduce weighted conformal prediction that relaxes the assumption of exchangeability to \textit{weighed exchangeability}.
This allows conformal prediction to be used in instances where there are distributional shifts between train and test samples, covariate shifts, i.e. $(X_i, Y_i) \sim P = P_X \times P_{Y|X}, i=1, \dots, n$ and $(X_{n+1}, Y_{n+1}) \sim \tilde{P} = \tilde{P}_X \times P_{Y|X}$, independently. 
It is important to note that while the distribution of $X$ changes, the conditional distribution $P_{Y|X}$ is assumed to be the same for training and test samples.

In this procedure, each nonconformity score $R_i$ is weighed by a probability proportional to the likelihood ratio  $w(X_i) = \frac{d\tilde{P}_X(X_i)}{dP_X(X_i)}$.\footnote{also called the Radon–Nikodym derivative.}
Due to this, instead of the empirical distribution of nonconfonformity scores which can be defined using multiple dirac delta distributions, a weighted version is defined as 
\[\sum\limits_{i=1}^{n} p_i^w(x) \delta_{R_i}+p_{n+1}^w(x)\delta_\infty, 
\]
where $\delta_{R_i}$ is the dirac delta, i.e., a point mass at point $R_i$.

The weights are defined by:
\[p_i^w(x) = \frac{w(x_i)}{\sum_{j=1}^{n}w(x_j) +w(x)},\; i=1, \dots, n.
\]
For a test instance, the weights are defined as:
\[
p_{n+1}^w(x) = \frac{w(x)}{\sum_{j=1}^{n} w(x_j) +w(x)}
\]
The prediction region is then given by:

\begin{multline}
    \widehat{C}_n(x) = \hat{f}(x) \pm \text{Quantile}\left(1-\alpha; \sum\limits_{i=1}^{n} p_i^w(x)\delta_{R_i}+p_{n+1}^w(x)\delta_\infty\right).
\end{multline}

\subsection{Treatment Effect Estimation}
Treatment effect estimation is a core concept in causal inference, widely applied in fields such as psychology, political science, and economics. 
The primary objective is to estimate the causal effect of a treatment (or intervention) on a target outcome variable.
To rigorously define causal effects, we adapt the potential outcomes framework originally introduced by \citet{neyman}, which provides a foundation for analyzing causal relationships in both observational studies and randomized experiments.

For each unit, let $Y_i(1)$ denote the potential outcome for an individual $i$ under treatment and $Y_i(0)$ denote the potential outcome for an individual $i$ under no treatment.
We can only observe one of those outcomes for each individual—this is known as the fundamental problem of causal inference. That is, we can never observe counterfactual outcomes. 
Given a sample of $n$ subjects and a binary treatment indicator  $T \in \{0, 1\}$, we define the following key quantities:

\begin{itemize}
    \item Average Treatment Effect: $ATE = \mathbb{E}[Y(1) - Y(0)]$
    \item Conditional Average Treatment Effect: $CATE = \mathbb{E}[Y(1) - Y(0)\; | \;X=x]$
    \item Individual Treatment Effect: $ITE = \tau_i = Y_i(1) - Y_i(0) \;\;\forall i\in n$
\end{itemize}
While in the past, research has focused on estimating the ATE and CATE, recently, because of the rise of personalized medicine, research has been trying to identify the ITE in observational and randomized studies.

In general, causal effect estimation relies on three key assumptions:

\begin{itemize}
    \item Stable Unit Treatment Value Assumption (SUTVA): Treatments are well defined and there is no interference between units, i.e. $Y = T*Y(1) + (1-T)Y(0)$. Specifically, there are no different levels or forms of the same treatment.
    \item Strong ignorability (Unconfoundedness): Conditional on observed covariates $X$, the treatment $T$ is independent of the binary outcome $Y(0)$ and $Y(1)$, i.e. $\{Y(0), Y(1)\} \perp T \;|\; X$.
    \item Overlap/Positivity: Each unit has a nonzero probability of receiving either treatment: $0 < \mathbb{P}(T=1\;|\;X=x) < 1$ for all $x$. This ensures comparable treatment and control groups.
\end{itemize}

Estimating treatment effects from observational data presents significant challenges due to a violation of assumption two, strong ignorability. 
A major issue is covariate shift, where the distributions of covariates differ between treated and untreated groups. 
Unlike randomized controlled trials (RCTs), where treatment assignment is independent of covariates, observational studies often assign treatments based on individual characteristics. 
This can distort the estimated effect of a treatment on an outcome when using conventional statistical methods.

Covariate shift is particularly problematic for estimating ITE and CATE because the distributional discrepancies between treatment and control groups are more pronounced at an individual level than when averaging over the entire population (as in ATE) \cite{machlanski_undersmoothing_2024}.
A common strategy to mitigate this issue involves reweighting techniques based on propensity scores \cite{athey_2019, Robins01091994}.
A propensity score is the conditional probability of a patient being assigned to a particular treatment given a set of covariates $X$, i.e. $e(x) =P(T=1|X=x)$
The method of propensity score matching uses these propensity scores by matching two individuals with similar treatments.
This method helps reduce bias and create more comparable groups. 
However, propensity score matching primarily improves treatment effect estimation within the common support of the covariates $X$, while it struggles with generalizing to points outside this region and is not individual-level, but group-level metrics.
Consequently, propensity score matching on its own is often not an adequate remedy for the covariate shift between treated and untreated groups - especially when trying to estimate conditional average treatment effects or individual treatment effects.
Moreover, it does not address model misspecification, which can significantly impact the accuracy of ITE estimates \cite{white}.

To address these challenges, model-agnostic meta-learners have been introduced \cite{metalearners}. 
These methods leverage various machine learning models (\textit{base learners}) — such as random forests, neural networks, and bayesian additive regression trees — to estimate key causal quantities, like outcome regressions or propensity scores. 
These meta-learners can be classified into two categories \cite{curth2021}: 
Indirect learners (one-step methods) and direct learners (two-step methods).
Indirect learners estimate regression functions separately for treated and untreated groups.
The difference between these two groups is computed and used to estimate the quantity of interest.
There are two notable examples of indirect learners: 
\begin{itemize}
	\item \textbf{S-Learner (Single-Model Learner):} Trains a single model 
	$\hat{f}(x,t) \approx \mathbb{E}[Y \mid X = x, T = t]$ 
	and estimates the treatment effect via 
	$\hat{\tau}(x) = \hat{f}(x,1) - \hat{f}(x,0)$.
	
	\item \textbf{T-Learner (Two-Model Learner):} Trains two separate models, 
	$\hat{f}_0(x) \approx \mathbb{E}[Y \mid X = x, T = 0]$ 
	and 
	$\hat{f}_1(x) \approx \mathbb{E}[Y \mid X = x, T = 1]$, 
	and estimates the treatment effect via 
	$\hat{\tau}(x) = \hat{f}_1(x) - \hat{f}_0(x)$.
\end{itemize}

Direct learners (two-step methods) on the other hand approximate the quantity of interest directly. 
Here, outcomes are residualized first, and thereby, the treatment effect can be estimated directly.
Notable examples include the following:
\begin{itemize}
	\item \textbf{X-Learner (Cross-Fitting Learner) \cite{curth2021}:} 
	First, estimate outcome models 
	$\hat{f}_0(x) \approx \mathbb{E}[Y(0)\mid X=x]$ and 
	$\hat{f}_1(x) \approx \mathbb{E}[Y(1)\mid X=x]$. 
	Next, construct pseudo-outcomes from the observed samples: 
	\[
	\tilde{d}_i^1 = y_i - \hat{f}_0(x_i) \quad \text{for units with } t_i = 1,
	\]
	\[
	\tilde{d}_i^0 = \hat{f}_1(x_i) - y_i \quad \text{for units with } t_i = 0.
	\]
	Then, fit two separate models $\hat{\tau}_0(x)$ and $\hat{\tau}_1(x)$ on these pseudo-outcomes, 
	using the subsets with $t_i=0$ and $t_i=1$, respectively. 
	Finally, define the CATE estimator as 
	\[
	\hat{\tau}(x) = g(x)\,\hat{\tau}_0(x) + (1-g(x))\,\hat{\tau}_1(x),
	\]
	where $g(x)$ is a weighting function, often chosen as the estimated propensity score $\hat e(x)$.

	\item \textbf{DR-Learner (Doubly Robust Learner) \cite{kennedy2023}:} 
	First, estimate the propensity score 
	$\hat{e}(x) \approx \mathbb{P}(T=1 \mid X = x)$. 
	Then, fit outcome models $\hat{f}_0(x) \approx \mathbb{E}[Y \mid X = x, T=0]$ 
	and $\hat{f}_1(x) \approx \mathbb{E}[Y \mid X = x, T=1]$. 
	Next, construct pseudo-outcomes for each observation $(x_i,t_i,y_i)$:
	\[
	\tilde{y}_i = \bigl(y_i - \hat{f}_{t_i}(x_i)\bigr)\frac{t_i - \hat{e}(x_i)}{\hat{e}(x_i)(1-\hat{e}(x_i))}
	+ \hat{f}_1(x_i) - \hat{f}_0(x_i).
	\]
	Finally, train a regression model on $\{(x_i, \tilde{y}_i)\}$ (often using a held-out sample) 
	to estimate the conditional treatment effect.
	
\end{itemize}

By leveraging machine learning techniques, meta-learners provide flexible and scalable approaches for estimating heterogeneous treatment effects, particularly in high-dimensional or nonparametric settings. 
However, challenges such as model selection, handling covariate shift, ensuring robustness to all forms of misspecification, and especially quantifying the uncertainty of point predictions of causal inference systems remain active areas of research.

\section{Review Methodology}
\label{sec:methodology}
A mapping review serves to systematically identify, categorize, and analyze existing literature on a given topic, with the objectives of classifying research contributions and identifying publication trends or patterns. 
This methodological approach was selected for our study as it enables both a comprehensive overview of current research developments and the identification of potential avenues for future investigation.
For any systematic review, research questions are of utmost importance in order to set focus \cite{10.5555/2349018}.
In the following, we define our research questions for this review.

\begin{itemize}
    \item RQ1: What are state-of-the-art methods using \textit{conformal inference} methods for \textit{treatment effect estimation}? 
    \item RQ2: What kind of \textit{treatment effects} have they been adapted to?
    \item RQ3: Which \textit{assumptions} do studies make?
    \item RQ4: Which \textit{models} are being used in case the methods are not fully model-agnostic?
\end{itemize}

Our methodological approach maintains strict alignment with the research questions throughout both the study selection process (see \autoref{study_selection}) and the presentation of the results (see \autoref{sec:results}). 
This consistent orientation ensures methodological coherence while enabling both comprehensive analysis and focused interpretation of findings. 
The research questions consequently serve as the conceptual foundation that systematically guides our investigation and subsequent knowledge synthesis.

\subsection{Search keyword construction}
\label{search_keyword}
Following the guidelines proposed by \citet{kitchenham_guidelines_2007}, we first decompose our research questions into fundamental search units comprising abbreviations, keywords, and phrases, which we then systematically combine using Boolean logic operations. Our search strategy employs a PICOC-like framework (Population, Intervention, Comparison, Outcome, Context) to structure these search units, where:
\begin{itemize}
\item \textit{Population}: Denotes specific roles or industry sectors.
\item \textit{Intervention}: Represents methodological approaches to problem-solving.
\item \textit{Comparison}: Indicates evaluations between alternative methodologies.
\item \textit{Outcome}: Encompasses the empirical results produced by interventions.
\item \textit{Context}: Defines the situational framework for these comparisons.
\end{itemize}

To tailor this framework to our specific review objectives, we introduce an extension by incorporating a second intervention term. 
This adaptation ensures our literature selection criteria exclusively capture studies that simultaneously employ conformal prediction methods as well as address treatment effect estimation.

\begin{itemize}
    \item Population: individual, patient, subject
    \item Intervention: treatment, ITE, CATE, counterfactual
    \item Second Intervention: conformal
    \item Comparison: learner, machine learning, model
    \item Outcome: interval, quantification, validity, robustness, counterfactual, outcome
    \item Context:  observation, trial, heterogeneous, randomized, empirical
\end{itemize}

Intra-search units are connected via the boolean operator \textit{OR} while inter-search units are connected using the boolean operator \textit{AND}.
Wildcard modifiers are used whenever possible and allowed by the search engine of the information source.

\subsection{Information source}
\label{information_source}
Information sources are an integral part of the identification process.
We use seven different electronic databases to search for relevant studies.

\begin{table}
    \centering
    \caption{Electronic databases used to search for relevant articles.}
    \label{tab:databases}
    \resizebox{\columnwidth}{!}{
    \begin{tabular}{ll}
        \toprule
        \textbf{Electronic Database}             & \textbf{Website} \\
        \midrule
        \emph{IEEE Xplore}                 & https://ieeexplore.ieee.org/Xplore/home.jsp            \\
        \emph{ACM Digital Library The ACM Guide to Computing Literature}             & https://dl.acm.org/               \\
        \emph{Wiley Online Library}             & https://onlinelibrary.wiley.com/            \\
        \emph{Science Direct} & https://www.sciencedirect.com/           \\
        \emph{Springer Link} & https://link.springer.com/ \\
        \emph{Taylor and Francis} & https://www.tandfonline.com//           \\
        \emph{Google Scholar} & https://scholar.google.com/           \\
        \bottomrule
    \end{tabular}
    }
\end{table}
The electronic databases are shown in \autoref{tab:databases}.
We use six primary databases and one meta-search engine with Google Scholar to identify relevant literature.
Because the search engines of each of these databases have limitations regarding the number of boolean operators, search terms, and much more, we must adjust the search terms for some of the information sources individually.
Science Direct and Google Scholar are affected by this adjustment and we note the adjustment in \autoref{sec:appendix}.

Regardless of the information source used, we restrict our search space to the years 2005-2025 since this research was conducted in early 2025.
2015 is used as the start year because conformal prediction has gained only recently from the statistics community attention.
In the end, $1250$ possibly related papers were found through our search in the digital databases using our aforementioned constraints.

\subsection{Study Selection}
\label{study_selection}
The selection of the most important papers from the $1250$ possibly related literature without missing any relevant paper is of utmost importance.
Consequently, we predefined a structured filtration process to find related literature:

\begin{enumerate}
    \item Remove impurities from our original search results. The databases might include brief reports, letters to the editor, etc. These are manually removed.
    \item Filter studies by F1, F2, F3, F4, and F5 by using the \textit{title} to remove irrelevant studies.
    \item Filter studies by F1, F2, F3, F4, and F5 by using the \textit{abstract} to remove irrelevant studies.
    \item Combine the search results.
    \item Remove all duplicated studies.
    \item Filter the studies by using the full text using filter criteria F1, F2, F3, F4, and F5.
    \item Filter the studies by using the full text using the quality criteria Q1 to Q9.
\end{enumerate}

To filter through the possibly related papers found in the electronic databases we use filtering statements, denoted with the variable $F$.
By the usage of these statements, only relevant papers are included and the filtration process is less subjective and reproducible.
We defined the filtration criteria as follows:

\begin{itemize}
    \item F1: Shall use conformal inference methods.
    \item F2: Shall be about treatment effect estimation.
    \item F3: No review papers shall be included.
    \item F4: No pure application paper shall be included.
    \item F5: Resultant output of methods shall be prediction regions.
\end{itemize}

Studies are not only filtered based on the content but also on the quality.
After the studies have been filtered with filtration process steps one to five, we filter by using the full text with the aforementioned filters but also with nine quality assessment criteria.
These criteria are essential to only include high-quality papers:

\begin{itemize}
    \item Q1: Is there a legible description of the research purpose?
    \item Q2: Is there an adequate description of the research context?
    \item Q3: Is there a review of related work?
    \item Q4: Is there a description of the conformal prediction method used?
    \item Q5: Is there a conclusion related to the research purpose?
    \item Q6: Is there a legible description of the application value?
    \item Q7: Is there novelty in the method being proposed?
    \item Q8: Does the study provide research orientation for further studies?
    \item Q9: Is there a description of the limitations of the method introduced?
\end{itemize}

If there are at least five of the nine quality criteria and if all of the filtration criteria F1-F5 are met, we include the paper in our study.

\begin{figure}
    \centering
    \includegraphics[width=\linewidth]{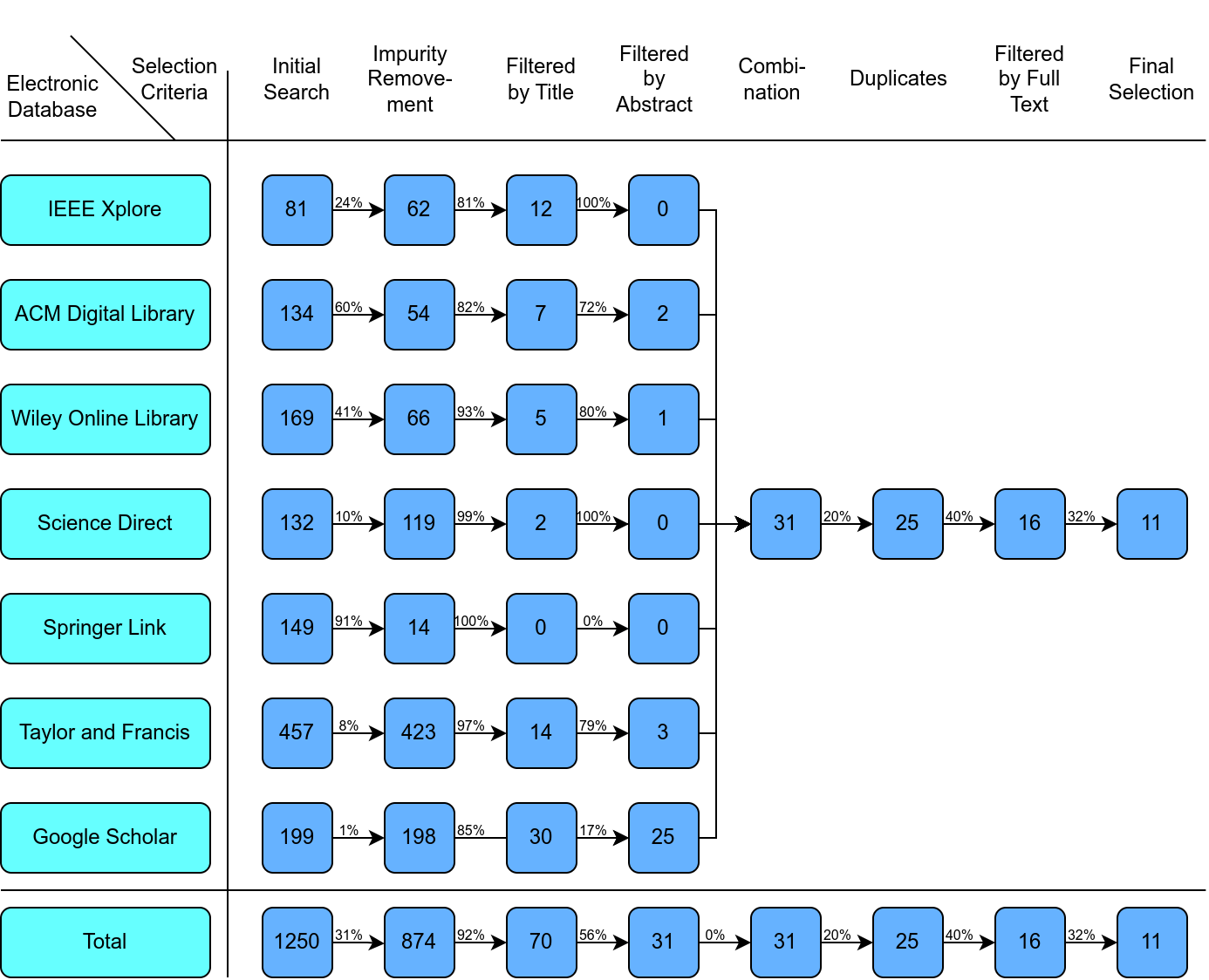}
    \caption{Literature Filtration Process.}
    \label{fig:papers}
\end{figure}

After filtering through the full text, we find that one of the papers was highly analogous to another in which the method is mostly the same, only the derivation differs.
In this case, we include the first paper.
The whole filtering process can be seen in \autoref{fig:papers}.

\begin{table*}
    \centering
    \caption{Overview of studies.}
    \label{tab:overview}
    \begin{tabular}{llll}
        \toprule
        \textbf{ID} & \textbf{Author}             & \textbf{Title} & \textbf{Year} \\
        \midrule
        1 & \citet{lei2021conformalinferencecounterfactualsindividual} & Conformal Inference of Counterfactuals and Individual Treatment Effects & 2021\\
        2 & \citet{jin_sensitivity_2022} & Sensitivity Analysis of Individual Treatment Effects: A Robust Conformal Inference Approach & 2023\\
        3 & \citet{zhang_conformal_2023}  & Conformal Off-Policy Prediction   & 2023\\
        4 & \citet{alaa_conformal_2024} & Conformal meta-learners for predictive inference of individual treatment effects & 2024 \\
        5 & \citet{cai_conformal_2024}& Conformal Diffusion Models for Individual Treatment Effect Estimation and Inference & 2024 \\
        6 & \citet{chen2024} & Conformal Counterfactual Inference under Hidden Confounding & 2024 \\
        7 & \citet{jonkers_conformal_2024} & Conformal Convolution and Monte Carlo Meta-learners for Predictive Inference of ITEs & 2024 \\
        8 & \citet{schroder_conformal_2024} & Conformal Prediction for Causal Effects of Continuous Treatments  & 2024\\
        9 & \citet{verhaeghe_conformal_2024} & Conformal Prediction for Dose-Response Models with Continuous Treatments & 2024 \\
        10 & \citet{gao_role_2025} & On the Role of Surrogates in Conformal Inference of Individual Causal Effects & 2025 \\
        11 &  \citet{wang_conformal_2025} & Conformal Inference of Individual Treatment Effects Using Conditional Density Estimates & 2025 \\
        \bottomrule  
    \end{tabular}
\end{table*}

We list the resultant studies chronologically in \autoref{tab:overview}.
There are a total of eleven studies that satisfy our filtration process and we describe them more in detail in the following.

\section{Results}
\label{sec:results_total}
\subsection{Conformal Prediction for Treatment Effect Estimation}
\label{sec:results}
The application of conformal prediction methods for treatment effect estimation presents several theoretical and practical challenges.
First, intervening on treatments induces a covariate shift in the covariate distribution, specifically, in the propensity score which violates the exchangeability assumption of standard conformal prediction methods.
Consequently, naive conformal prediction intervals for treatment effects become marginally invalid and lose their finite-sample coverage guarantees.

This necessitates methodological adaptations that explicitly account for distributional differences between treated and untreated units.
However, this assessment requires information about the propensity score which is typically unknown in observational studies.
The need to estimate propensity scores introduces additional uncertainty that must be properly incorporated into the interval construction process to maintain validity—a core challenge for observational studies since a selection bias often exists. 

Moreover, even standard assumptions (e.g., no unmeasured confounding) are often violated in observational studies but researchers and practitioners still require prediction intervals that are both sharp and valid. 
In the following sections, we survey recent methodological innovations addressing these fundamental challenges.
\newline

\noindent 
The foundational work by \citet{lei2021conformalinferencecounterfactualsindividual} introduced the first conformal prediction approach for individual treatment effects, extending the weighted split conformal quantile regression framework originally developed by \citet{tibshirani2020conformalpredictioncovariateshift}. 
Their method produces reliable interval estimates for counterfactuals and individual treatment effects where the intervals have guaranteed marginal coverage in finite samples under completely randomized experiments.
Even for randomized trials with imperfect compliance or for observational studies when the propensity score $e(x)$ is unknown and needs to be estimated, marginal coverage is approximately guaranteed if the estimate of the propensity score $\hat{e}(x) \approx e(x)$ or the conditional quantiles $\hat{q}_\beta(x) \approx q_\beta(x)$ with $\beta \in {\alpha_{lo}, \alpha_{hi}}$, where $q_\beta(x)$ is the $\beta$-th quantile of Y(1) (or Y(0)) given $X = x$ can be accurately estimated.
This assumes strong ignorability in data, SUTVA, as well as positive overlap and implies that either the outcome model or the treatment model is accurately estimated.
The constructed intervals thus satisfy a doubly robust property since either the estimation of the propensity scores or the estimation of the quantiles needs to be accurate.
If the conditional quantiles are estimated accurately, then the weighted split CQR procedure even has approximately guaranteed conditional coverage.

\citet{lei2021conformalinferencecounterfactualsindividual} first perform counterfactual inference to generate valid intervals for subjects in the studies.
In order to perform counterfactual inference in observational data, i.e. measuring the treatment effect of the counterfactual outcome for subjects in the study, the weights for the treated and untreated population depend majorly on the propensity score, s.t. $w_0(x)=\hat{e}(x)/(1-\hat{e}(x))$ and for the treated population $w_1(x)=(1-\hat{e}(x))/\hat{e}(x)$.
For fully randomized experiments, unweighted conformal prediction may be used.

Additionally, \citet{lei2021conformalinferencecounterfactualsindividual} extend their approach from estimating intervals of counterfactuals for subjects in the study to ITEs for subjects not included in the study.
They create a pair of prediction intervals at level $\frac{1-\alpha}{2}$, i.e., $[\hat{Y}^L(1;x)-\hat{Y}^R(1;x)]$ for $Y(1)$ and $[\hat{Y}^L(0;x)-\hat{Y}^R(0;x)]$ for $Y(0)$, where $\hat{Y}^L$ and $\hat{Y}^R$ denote the predicted lower and upper prediction bound.
In this naive approach, the intervals of the two estimated intervals are contrasted and created by $\hat{C}_{ITE}(x)=[\hat{Y}_L(x,1)-\hat{Y}_R(x,0), \hat{Y}_L(x,1)-\hat{Y}_R(x,0)]$.
If the estimated counterfactual intervals have guaranteed coverage, then the estimated ITE intervals will also have guaranteed coverage.
They also provide a nested approach that relies on splitting the data into two folds by first training to get the prediction intervals $\hat{C}_1(x)$ and $\hat{C}_0(x)$ on the first fold.
Thereafter, they compute for each unit in the second fold $\hat{C}_0(X_i)$ if $T_i=1$ and vice versa. 
This induces a surrogate interval 
\[
\hat{C}_{ITE}(x, t, y^{obs})=
\begin{cases}
y^{obs}- \hat{C}_0(x), t=1, \\
\hat{C}_1(x)-y^{obs},t=0.
\end{cases}
\]
In the exact version, for any testing point, they then apply unweighted conformal inference by merging both folds on the generated counterfactual intervals with a model that estimates the conditional mean (or median).
This estimation is used for the nonconformity score.
In the inexact version, they simply fit conditional quantiles of the lower and upper prediction regions in the two folds.
In empirical experiments, they show that the inexact version often achieves similar coverage as the two other versions while the intervals are much more narrow.

\subsubsection{Meta-Learners}
\citet{alaa_conformal_2024} apply standard conformal prediction procedures on top of a wide variety of meta-learners using two-stage pseudo-outcome regression.
For the construction of valid intervals, they assume the commonly used assumptions of SUTVA and strong ignorability, as well as that the propensity scores are known.
Because they use two-stage meta-learners, they first estimate pseudo-outcomes and then regress those pseudo-outcomes on covariates to obtain point estimates of CATE.
Thereafter, intervals for the individual treatment effect are constructed by computing the empirical quantile of conformity scores evaluated on the pseudo outcomes on a holdout calibration set. 
Because these intervals are pseudo-intervals constructed on pseudo-outcomes, it would be highly likely that these intervals are marginally invalid.
They prove, however, that the procedure is valid (i.e. has marginal coverage) if their conformity scores of the pseudo-outcomes stochastically dominate the real ("oracle") conformity scores.
Specifically, they show, that if either the conformity scores of the meta-learners are stochastically larger (first-order stochastic dominance) or if they have a larger spread (second-order stochastic dominance, then the prediction intervals are valid for high-probability coverages. 
They show that this is the case for the DR-Learner as well as the IPW-Learner.
In cases when conformity scores more strongly dominate oracle scores, the model performs worse, and interval widths are wide, however.
\newline

\noindent 
\citet{jonkers_conformal_2024} introduce two novel approaches: 
A method called conformal convolution T-learner (CCT-Learner) uses convolution, conformal prediction, and also propensity score weighting to address covariate shifts.
A second method called conformal Monte Carlo meta-learners (CMC) uses Monte Carlo sampling to approximate predictive distributions losing finite-sample guarantees as well as performance in favor of computational efficiency.
Both of these methods make use of conformal predictive systems introduced by \citet{vovk2017}, which allow to derive full predictive distributions under i.i.d. assumptions.
Specifically, their approach uses a T-Learner and estimates the predictive distribution for both factual and counterfactual outcomes reweighted by propensity scores.
Both approaches use weighted conformal prediction to combat the covariate shift, thus they need access to the propensity scores that are assumed to be known.
For the CCT-Learner, after probabilistically calibrated predictive distributions of counterfactual outcomes have been generated, they perform a convolution to infer the ITE.
For the CMC-Learner, they instead sample from each estimated potential outcome distribution and use those samples to produce ITE samples.
They also note the existence of the \textit{epsilon problem}:
The joint distribution of counterfactual noise terms $\epsilon(0)$ and $\epsilon(1)$ is unidentifiable.
Thus, violations of independence lead to compromised distributional validity.
This is implicitly also assumed in other methods and they show the degradation of the coverage if independence is violated empirically.

\subsubsection{Sensitivity Model}
\citet{jin_sensitivity_2022} propose a model-free framework for assessing the robustness of individual treatment effects against potential confounding factors using sensitivity analysis.
This is done via the usage of conformal prediction to provide valid prediction intervals for counterfactual outcomes.
The usage of such sensitivity models aids researchers in understanding how sensitive their causal analysis is to potentially unmeasured confounding factors.
Their method achieves marginal coverage when the propensity score is known and approximate marginal coverage when it is approximately estimated.
The authors use Rosenbaum's Marginal Sensitivity Model (MSM) to quantify the impact of unmeasured confounding.
The key parameter is $\Gamma$, which measures how much the true treatment assignment probability differs from the observed propensity score $e(x)$.
To incorporate sensitivity analysis, the authors introduce a robust weighted conformal inference framework in which each unit is weighted based on the worst-case confounding allowed under a given $\Gamma$.
Thus, even if there is hidden confounding, the procedure remains valid. 
However, their method needs access to upper bound $\hat{u}(x)$ and lower bound $\hat{l}(x)$  of the likelihood ratio $w(x,y)$ between the observational and interventional distribution in order to characterize the covariate shift.
They state that even if the estimated bounds are large, the procedure remains valid.
According to their experiments, neither the estimation of $\hat{e}(x)$, nor $\hat{l}(x)$ and $\hat{u}(x)$ needs to be accurate, but instead it matters that $\hat{l}(x)$ is below $w(x,y)$ and $\hat{u}(x)$ is above it. 
By creating valid prediction regions, the authors test the hypothesis that the true value is included in a given sample.
Either the true value is not included because the individual treatment effect was really not in the prediction region or the confounding level was at least $\Gamma$.
Based on the proportion of such rejections for different levels of confounding, the authors then reject the hypothesis that a confounding level is at least a certain $\Gamma$. 
\newline

\noindent
\citet{chen2024} introduce a conformal prediction method that does not require the strong ignorability assumption, i.e. their method can be used even in the presence of hidden confounders (in contrast to e.g. \citet{alaa_conformal_2024, lei2021conformalinferencecounterfactualsindividual} and does not need the aforementioned upper and lower bounds in contrast to \citet{jin_sensitivity_2022}).

However, their method still requires evaluating the distributional shift.
Thus, they not only use observational data but also interventional data to account for the covariate shift between treated and untreated units.
This data need thus to be taken from randomized control trials and their method, weighted conformal prediction with density ratio estimation, can only be applied if such data is available.
Their method estimates the density ratio between the interventional and observational distribution using randomized control trial data, thus, they can provide coverage guarantees even under the presence of confounding.

\subsubsection{Continuous Treatments}
\citet{schroder_conformal_2024} introduce a novel method for estimating prediction intervals for potential outcomes in continuous treatment settings.
They create prediction intervals that are valid even if the propensity score is unknown and needs to be estimated from the data at hand.
The prediction intervals can be wide, however, if the propensities are estimated poorly.
Additionally, prediction intervals are conservative for the point intervention or if calibration data is scarce, meaning that a representative and large calibration data set is essential for narrow prediction intervals. 
Lastly, their approach is computationally expensive since it requires solving a non-convex optimization problem (e.g. interior point methods) per confidence level, treatment level, and sample whenever they want to provide a prediction interval.
\newline

\noindent
\citet{verhaeghe_conformal_2024} tackle the issue of creating valid prediction intervals for dose-response models with continuous treatments by combining those with weighted conformal prediction to account for the covariate shift in observational studies.
They use a conditional average dose-response function model trained on triples $(X,T,Y)$ which is used for querying the dose-response for all allowed treatment levels creating an interventional distribution.
The proposed weights are kernel functions such that for each new sample $x_0$ a new calibration procedure must be performed and they are dependent on not only the sample but also on the target treatment $t$.
This ensures coverage around $x_0$ and allows for dynamic intervals around that point providing a heteroskedastic approach.
In their study, however, the target interventional treatment distribution is assumed to be known beforehand and thus can be computed in advance.
A notable point is that treatment levels between lower treatment level $t_L$ and upper treatment level $t_U$ are all assumed to be equally likely, i.e. they assume a uniform distribution between $t_L$ and $t_U$ which might not be true for all applications.

This is distinct from the method of \citet{schroder_conformal_2024} as they are concerned about (soft and hard) treatment interventions rather than providing prediction intervals for dose-response models. 
This means, that while \citet{verhaeghe_conformal_2024} are concerned about providing a dose-response model (requiring a uniform assumption about treatment levels) \citet{schroder_conformal_2024} are concerned about quantifying the causal effect a single intervention has. 
\newline

\subsubsection{Generative Models}
\citet{cai_conformal_2024} propose a novel conformal diffusion model for treatment effect estimation in observational studies. 
Their method uses two deep generative models to learn the conditional distribution of a potential outcome $P(Y(T)\;|\;X)$ for a binary treatment. 
Thereafter, they compute the mean from random samples of the learned distributions and create interval estimates of the random samples.
They adjust for two possible distributional shifts: A covariate distributional shift occurs because the joint distribution for treated subjects is $P_{X|T=1} \times P_{Y(1)|X}$ while for test data the corresponding joint distribution might differ, i.e., $Q_X \times P_{Y(1)|X}$, where $Q_X$ is the covariate distribution in the test population; and a distributional shift occurring because in observational studies, usually, $P_{X|T=0} \neq P_{X|T=1}$. 
The covariate distributional shift between the two generative models is accommodated using a local approximation for covariates, measuring the covariate similarity between the calibration data and the target one and then reweighing the nonconformity scores based on this similarity. 
The second potential distributional shift between treated and control groups is adjusted by using the classical propensity score adjustment.
Thus, their method relies on a proper estimation of the real propensity scores as well as the approximation error of the kernel function.
If both of those errors go to zero as the size of calibration data and test data goes to infinity, then their method has asymptotic coverage.

\subsubsection{Narrowing prediction intervals}
\citet{gao_role_2025} integrate so-called surrogates, variables that act like proxies for unobserved or hard-to-measure outcomes, to enhance the calibration of prediction intervals for individual treatment effects.
The surrogates provide additional information thereby improving the efficiency of the calculated prediction intervals.
The use of surrogates imposes some additional assumptions on the procedure, specifically the surrogates themselves.
In addition to the three assumptions of causal inference, \citet{gao_role_2025} require that the surrogates are conditionally independent of treatment and that surrogates are distinctly informative for the potential outcomes.
Uninformative surrogates could widen intervals.
Moreover, their method only provides probably approximately correct asymptotic coverage though their approach achieves group-conditional coverage over previously specified subgroups.
\newline

\noindent
\citet{wang_conformal_2025} propose a method to calculate prediction intervals that are marginally valid and narrower than existing methods theoretically and experimentally.
They adopt the two-stage framework by \citet{lei_conformal_2021} but use conditional density as a nonconformity score to decrease the prediction interval lengths while ensuring the same desired coverage guarantees.
Estimating the conditional density presents significant challenges.
\citet{wang_conformal_2025} use a reference distribution technique and recommend the usage of a Gaussian distribution.
This conditional density estimate is then used as a (non)conformity measure in the weighted conformal prediction procedure.

\subsubsection{Off-Policy Evaluation}
The off-policy evaluation framework (OPE) usually focuses on a contextual bandit setting in which there are observable triplets $\{(X_i, T_i, Y_i)\}^n_{i=1}$, where $X_i$ is defined as the contextual information of the $i$-th instance, $T_i \in \{0, 1, \dots, m-1\}$ denotes the treatment or action, and $Y_i$ is the corresponding response (outcome or also called reward).
$Y_i^t$ denotes the reward of the $i$-th instance if they received action $t$.
A policy $\pi$ is a stochastic decision rule that maps the contextual distribution function over the action space.
$\pi(t|x)$ is the probability that the agent selects treatment given $X=x$.
$\pi_e$ denotes the target policy while $\pi_b$ denotes the behavioral policy.
Additionally, the standard causal effect assumptions are imposed as defined earlier.

They provide three methods for conformal prediction to the OPE problem, a direct one, which uses weighed conformal prediction directly since the calibration dataset and the outcome pair $(X_{n+1}, Y_{n+1}^{\pi_e})$ is weighed exchangeable.
However, for this $w_{n+1}= \frac{dP_{Y^{\pi_e}|X}(y|x)}{dP_{Y|X}(y|x)}$ needs to be estimated which is highly challenging.
The authors provide a second method, the subsampling method, in which the distributional shift is handled by taking a subset of the data whose distribution is similar to the target distribution and for which conformal prediction is applied.
In particular, they sample a pseudo action $E_i$ following the evaluation policy $\pi_e$, select subsamples that match the observed actions, and apply conformal prediction to the subsamples.
This approach is not valid and prediction intervals might undercover or overcover because the distribution of the selected response differs from that of the potential outcome.

Their proposed method alleviates this issue by creating an auxiliary policy $\pi_a$ whose distribution depends on both the target policy $\pi_e$ and the behavioral policy $\pi_b$ such that its conditional distribution is the same as $P_{Y^{\pi_e}|X}$ in the target population.
Then, still, the covariate distribution is different between target and subsamples but this can be handled by weighted conformal prediction.
The weight $w_{n+1}$ depends on the behavioral policy which only is known in randomized studies, otherwise, it needs to be estimated with the usage of supervised learning.

The authors also extend their method based on importance sampling and multi-sampling and generalize their approach for sequential decision-making.

\subsection{Meta-Data Analysis}
In this review, we also discuss and analyze the meta-data of the publications, i.e., when and where papers were published.

\begin{figure}
    \centering
    \includegraphics[width=\linewidth]{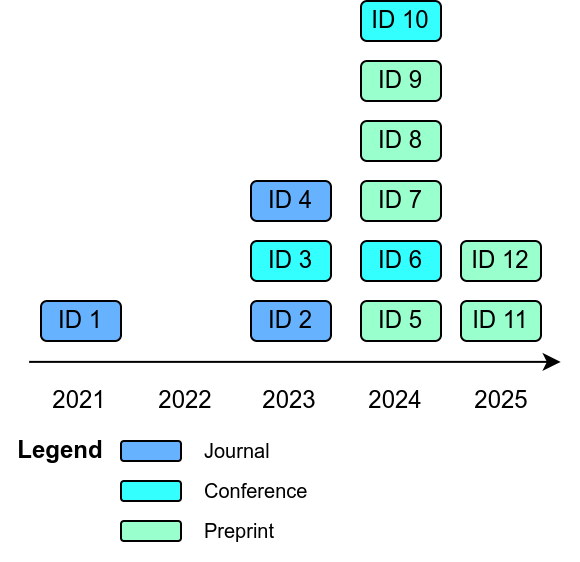}
    \caption{Literature Filtration Process.}
    \label{fig:publication_chronology}
\end{figure}

For this, we show the publication chronology in \autoref{fig:publication_chronology}.
As can be seen, all papers have been published between 2021 to 2025, with an increasing trend in the more recent years.
Thus, methods have been only developed in recent years, arguably because of the recent advancements in conformal prediction by making it computationally inexpensive (cf. \citet{splitcp} and \citet{splitcplei}), allowing it to be used in settings where distribution shifts happen (cf. \citet{tibshirani2020conformalpredictioncovariateshift}), and by tightening prediction regions with advanced techniques (cf. \citet{romano2019conformalizedquantileregression}).

Even though the amount of papers in this area is still relatively small, because of the data-driven, distribution-free nature and the strong theoretical guarantees of conformal prediction, more papers might be published in the next few years.

Besides, all studies identified through the filtration process are in one of three categories:
Journal papers, conference papers, or preprint papers.
Preprints are the most common publishing type, mostly published in the years 2024 and 2025.
It is likely that these papers have been submitted at conferences or journals but are still under review.
Some of the papers that have been published and accepted by a scientific journal or conference (e.g. \cite{lei_conformal_2021, jin_sensitivity_2022, alaa_conformal_2024} have been published at high-ranking publishers like the Journal of the Royal Statistical Society Series B, Proceedings of the National Academy of Sciences or Advances in Neural Information Processing Systems.
This emphasizes the importance of this research field.

\section{Discussion}
\label{sec:discussion}

\subsection{Future Work}
This review paper identified key literature for researchers entering this promising and young research field.
Moreover, we also provide promising research directions for future work.

Most research in \autoref{sec:results} focused on applying weighted (quantile) conformal prediction to a number of tasks characterized by varying application scenarios, including binary outcomes, continuous outcomes, and the use of the off-policy evaluation framework.
The main question they were concerned with is how to adjust the conformal prediction procedure in a way such that the procedure gives valid guarantees even though there are distributional differences between treated and untreated subjects through unknown propensity scores which necessitates the estimation of those propensity scores in observational studies.
Furthermore, they differed in assumptions about data or only offered asymptotic or approximate guarantees.

Because of the inherent distributional difference between treated and untreated subjects, it will always be relevant how weights can be chosen in a way to guarantee (or approximate) the validity of the constructed intervals.
For this, the distributional change of treated and untreated subjects would need to be known (or approximated).
It would be valuable to have this validity without having to rely on some part of the interventional distribution (cf. \citet{chen2024}) since in many areas, there is no data from such distribution available because of e.g. ethical concerns.
\citet{auer_conformal_2023} have been using weighted conformal prediction in conjunction with Modern Hopfield Networks (MHN) and have successfully used the associative memory of these MHNs to find more similar examples in the past for time series data.
This allowed \citet{auer_conformal_2023} to find better weights for the conformal prediction procedure which led to higher validity and tighter intervals.

Another important recurring theme in many papers is that intervals are overly conservative to be practical for real applications, i.e. the width of the intervals is too wide.
Additionally, the produced intervals are often static, i.e. not adapting to the difficulty of the individual subject at hand.
Thus, no local variability in the uncertainty of a prediction is incorporated.
This concept is usually formalized through the idea of \textit{adaptivity} (cf. \autoref{sec:preliminaries}), i.e. achieving conditional coverage.
While achieving conditional coverage is impossible, research efforts should focus on creating individualized, adaptive intervals because the width of the prediction region only majorly depends on the estimator's performance, the nonconformity measure, and the conformal prediction technique.
Thus, adapting nonconformity measures to include the local variability of a sample is an important research direction that has been applied to other settings in conformal prediction before.

\subsection{Limitations}
While we ensured a structured and transparent process in how we searched and filtered to find relevant research papers, we need to mention the limitations thereof.
There are three different limitations.

\begin{enumerate}
    \item Bias in the search strategy.
    \item Bias in applying filter criteria.
    \item Bias by limiting papers in Google Scholar search.
\end{enumerate}

We employ a specific search strategy in order to minimize any bias in the literature search process.
Still, especially the keyword construction is by definition subjective which may lead to important literature missing, or more generally, to not fully comprehensive search results.

Additionally, while we were using seven electronic databases in which we found 1250 papers with our search strategy, there might be other electronic databases that are not covered and potentially have interesting papers concerning this topic.
We described the whole process extensively in \autoref{sec:preliminaries}.

Finally, we define seven quality assessments that ensure that the survey process is unbiased.
Again, however, the creation of such quality assessments is subjective and the application of these classification criteria to papers is also highly subjective.

\section{Conclusion}
\label{sec:conclusion}
\begin{table*}
	\centering
	\caption{Assumptions of identified papers.}
	\label{tab:assumptions}
	\begin{tabularx}{\linewidth}{llllXX}
		\toprule
		\textbf{Paper} & \textbf{SUTVA} & \textbf{Strong Ignorability} & \textbf{Overlap} & \textbf{Additional Assumptions} \\
		\midrule
		\citet{lei2021conformalinferencecounterfactualsindividual} & \cmark & \cmark & \cmark & Accurate estimation of propensity scores or conditional quantiles of potential outcomes\\
		\citet{jin_sensitivity_2022} & \cmark  & \xmark & \cmark & Accurate estimation of lower and upper bound for likelihood ratio\\
		\citet{zhang_conformal_2023} & \cmark & \cmark & \cmark & Sequential ignorability\\
		\citet{alaa_conformal_2024} & \cmark & \cmark & \cmark & Propensity score is known\\
		\citet{cai_conformal_2024} &  \cmark & \cmark & \cmark & Accurate estimation of propensity scores and approximation error of kernel function \\
		\citet{chen2024} & \cmark & \xmark & \cmark & Access to fraction of interventional distribution\\
		\citet{jonkers_conformal_2024} & \cmark & \cmark & \cmark & Independent potential outcome noise distributions and knowledge of propensity score \\
		\citet{schroder_conformal_2024} & \cmark & \cmark  & \cmark & Estimation error of propensity score bounded \\
		\citet{verhaeghe_conformal_2024} & \cmark & \cmark & \cmark & Treatment levels uniformly distributed and no distribution shift between interventional and observational distribution\\
		\citet{gao_role_2025} &  \cmark &  \cmark &  \cmark &  Additional assumptions on surrogates\\
		\citet{wang_conformal_2025} & \cmark & \cmark & \cmark & Accurate estimation of propensity score \\
		\bottomrule  
	\end{tabularx}
\end{table*}
In this paper, we conducted a review study on conformal prediction methods for estimating treatment effects in observational as well as partly or fully randomized studies.
We performed a systematic literature search including seven electronic databases and selected eleven studies guided by our filtration process and after careful examination.

In respect to the papers examined, we find that the paper by \citet{lei_conformal_2021} is a foundational paper which most papers build upon (cf. \cite{jin_sensitivity_2022, alaa_conformal_2024, cai_conformal_2024, gao_role_2025}) or compare to.
All papers rely on the concept of \textit{weighted exchangeability} to characterize the distribution shift, introduced by \citet{tibshirani2020conformalpredictioncovariateshift}.
This concept has been applied to different tasks, namely modeling the dose-response function, and off-policy evaluation, to name a few.
Others replaced some of the usual conformal inference assumptions with others.
We find with respect to our research questions defined in \autoref{sec:methodology}:

RQ1: 
State-of-the-art methods heavily rely on the data and tasks at hand. 
The performance of models is measured with two metrics: 
Average interval widths and achieved coverage.
The data is usually simulated and generated with error characteristics exhibiting heteroscedasticity and homoscedasticity.
Authors usually use the data generation process of \citet{wager2017estimationinferenceheterogeneoustreatment}.
Some semi-synthetic data sets are also popular.

RQ2: One paper is concerned with deriving bounds for off-policy prediction \citet{zhang_conformal_2023}.
Every other study derives bounds for the \textit{individual treatment effect}.
RQ3: We list all the assumptions of the individual studies in \autoref{tab:assumptions}.
Most of the studies assume the three necessary treatment effect assumptions.
Additionally, all studies assume that the individual subjects are i.i.d.

RQ4: All models are model-agnostic in the sense that any machine learning model can be used to make predictions for counterfactuals or the treatment effect. 
However, some methods assume some meta-models.  
\citet{jin_sensitivity_2022, verhaeghe_conformal_2024, alaa_conformal_2024} assume an MSM, a dose-response model, and DR/IPW-learner respectively.
\citet{lei2021conformalinferencecounterfactualsindividual, wang_conformal_2025, gao_role_2025, chen2024, jonkers_conformal_2024, schroder_conformal_2024} do not assume the usage of any specific models.

Overall, the current research on treatment effect estimation and conformal prediction has advanced significantly.
The recent introduction of weighted conformal prediction by \citet{tibshirani2020conformalpredictioncovariateshift} enabled the usage of conformal prediction for treatment effect estimation.
By this introduction, and the theoretical results by \citet{lei_conformal_2021}, conformal prediction has been successfully applied to the treatment effect estimation problem in observational and randomized studies in which propensity scores are unknown and known respectively.
Regardless, there are still significant challenges.
The assumptions for observational studies for treatment effect estimation are often not easy to justify, especially the strong ignorability assumption.
In the future, more efforts could be dedicated to tightening prediction intervals and working towards conditional (or group-conditional) coverage in practice.

\FloatBarrier

\bibliographystyle{ACM-Reference-Format}
\bibliography{references}

\appendix
\section{Appendix}
\label{sec:appendix}
For science direct, we had to restrict the search to fewer boolean operators, thus, we only used the search term \textit{(individual OR patient OR subject) AND (treatment OR ITE OR CATE) AND (conformal inference OR conformal prediction)} to find relevant literature.
For google scholar we are more restrictive with our search clause because google scholar is a meta-search engine which crawls through a number of different journals which makes the amount of search results too high for this literature review.  
Consequently, for the \textit{2nd Interventions}, instead of broadly specifying \textit{conformal}, we specify \textit{conformal inference OR conformal prediction} because otherwise too many search results were found.
We further limit the google scholar search to the first 199 relevant literature sources.

\end{document}